# Modified Soft Brood Crossover in Genetic Programming

Hardik M. Parekh

*Information Technology Department*
*Dharmsinh Desai University*
*Nadiad, India*
hrdk018@gmail.com

Vipul K. Dabhi

*Information Technology Department*
*Dharmsinh Desai University*
*Nadiad, India*
vipul.k.dabhi@gmail.com

*Abstract –* **Premature convergence is one of the important issues while using Genetic Programming for data modeling. It can be avoided by improving population diversity. Intelligent genetic operators can help to improve the population diversity. Crossover is an important operator in Genetic Programming. So, we have analyzed number of intelligent crossover operators and proposed an algorithm with the modification of soft brood crossover operator. It will help to improve the population diversity and reduce the premature convergence. We have performed experiments on three different symbolic regression problems. Then we made the performance comparison of our proposed crossover (Modified Soft Brood Crossover) with the existing soft brood crossover and subtree crossover operators.**

## I. INTRODUCTION

Genetic programming is a model of programming which uses the ideas of biological evolution to handle a complex problem. From the number of possible solutions, the most effective solutions survive and compete with other solutions in such a way to reach closer to the needed solution. Premature convergence is one of the most important issues while using Genetic Programming for data modeling. Premature convergence leads to evolution of solutions which are locally optimal. Premature convergence can be avoided by improving population diversity. Population diversity can be improved using intelligent crossover. Our research aim is to improve population diversity using intelligent crossover. We have analyzed different toolkits available for GP and found the JCLEC[3] (Java Class Library for Evolutionary Computation) useful for our research work. JCLEC is an open source, platform independent and implemented in java.

## II. INTELLIGENT CROSSOVER OPERATORS

Crossover is an important operator in genetic programming. Standard crossover may produce the offspring same as their parents. Standard crossover does not having intelligence that how to avoid this problem of generating offspring same as their parents. Intelligent crossover combines the parents in such a way that it can generate the offspring having better fitness than their parents. Premature convergence leads to evolution of solutions which are locally optimal. To evolve globally optimal solutions, avoidance of premature convergence is required. Our aim is to avoid premature convergence during GP run and hence we have to improve the population diversity. Intelligent crossover operator can be useful to improve population diversity. We have analyzed few intelligent crossover operators like Context Aware Crossover (CAC) [4], Semantic Aware Crossover (SAC) [1], Semantic Similarity based Crossover (SSC) [5], Soft Brood Crossover [2], Approximating Geometric Crossover [6], Selective Crossover [8] and Size Fair Crossover [7].

From the comparison of different crossover operators, bases on different criteria specified in Table 1, we have observed that soft brood crossover operator can be useful to improve the population diversity.

TABLE I
COMPARISON OF INTELLIGENT CROSSOVER OPERATORS

| | Context Aware X-over [4] | Semantic Aware Crossover [1] | Semantic Similarity based Crossover [5] | Soft Brood Crossover [2] | Approximating Geometric Crossover [6] | Size Fair Crossover [7] | Selective X-over [8] |
|---|---|---|---|---|---|---|---|
| 1. Crossover point selection | One subtree randomly selected | Randomly select the point | Randomly select crossover points in both trees | randomly | Randomly | First crossover point selected randomly Second one subtree size must not exceed to first one | Best node |
| 2. Swapping criteria | All possible subtree crossed over with first one | If two subtrees are not semantically equivalent then perform x-over | Find dissimilar subtrees until max trial attempt | Randomly N times crossover | N times crossover | Size similarity | Two best node (fitness) |
| 3. After crossover | Based on fitness best child selected. | Expected to change in semantics of whole trees. | Expected to change in semantics of whole trees | Best two fittest are selected | Semantically most similar to both parents | Best fittest are selected | Best fittest node |
| 4. Efficiency of crossover operator | fitness | Semantic diversity | Semantic diversity | fitness | Semantically geometric | Fitness | Fitness |

### A. Soft Brood Crossover(SBC)

Soft brood crossover differs from the other crossover operators. The number of crossover performs on the same pair of parents. The performed operations generate the number of offspring. Then each offspring is evaluated based on their fitness. From the generated offspring two best fittest offspring passes into next generation and the rest are discarded.

## B. Modified Soft Brood Crossover(MSBC)

We have proposed a crossover operator that modifies existing soft brood crossover operator. This can help to prevent premature convergence and improve the population diversity. Algorithm1 presents the pseudo code for proposed crossover operator in detail.

---

**Algorithm 1:** *Modified Soft Brood Crossover*

---

1. Select parent P1 and P2 for crossover
2. N random crossover operations are performed to generate a brood of 2N children
3. If generation <= (1/2) Total generation
   a. The fitness of all children is calculated
   b. Most two dissimilar (based on fitness) children are copied into next generation and rest are discarded
4. Else
   a. The evaluated children sorted based on their fitness
   b. Two best fittest children are copied into next generation and other are discarded.

---

Modified Soft Brood Crossover operator generates the number of offspring from the same pair of parent. For the first half of the generation we are passing two most dissimilar offspring into next generation based on their fitness. And for the rest half generation we are passing two best fittest offspring into next generation.

TABLE II
RESULTS FOR DIFFERENT RATIO OF GENERATIONS

| The ratio of Generation for the Modified Soft brood crossover | | Best fitness found at the generation for Equation1 | Best fitness found at the generation for Equation2 (sextic) | Best fitness found at the generation for Equation3 |
|---|---|---|---|---|
| Percentage of total generation operated in the first half | Percentage of total generation operated in second half | | | |
| 10% | 90% | 0.019478672 at gen-48 | 0.696866493 at gen-6 | 0.948267753 at gen-3 |
| 20% | 80% | 0.015173981 at gen-17 | 0.896934359 at gen-27 | 0.948267753 at gen-2 |
| 30% | 70% | 0.011116504 at gen-19 | 0.891920931 at gen-36 | 0.948267753 at gen-3 |
| 40% | 60% | 0.016501593 at gen-38 | 0.596869303 at gen-21 | 0.948267753 at gen-2 |
| 50% | 50% | 0.007081805 at gen-45 | 0.464426127 at gen-40 | 0.948267753 at gen-2 |
| 60% | 40% | 0.024519353 at gen-49 | 0.999961166 at gen-2 | 0.948267753 at gen-2 |
| 70% | 30% | 0.014934063 at gen-38 | 0.797835834 at gen-2 | 0.948267753 at gen-2 |
| 80% | 20% | 0.023891294 at gen-42 | 0.497219429 at gen-38 | 0.948267753 at gen-2 |
| 90% | 10% | 0.012435098 at gen-42 | 0.59264934 at gen-31 | 0.948267753 at gen-2 |

We performed 30 runs for each problem using different percentage of generations by passing them into the first and second half of the crossover operator. From the obtained results represented in Table II, we found that passing the 50% of generations in the first half and the 50%of generations in the second half gives the best results.

Problem 1: $cos\ (\sqrt{sin}\ (q))\ *\ cos\ (p)\ *\ sin(x)\ +\ tan(r\text{-}s)$
Problem 2: *Sextic Polynomial:* $x^6 - 2x^4 + x^2$
Problem 3: $2x^2\text{-}3x\ +4$

## III TOOLKIT

We have analyzed the different toolkits available for GP and found JCLEC[3] (Java Class Library for Evolutionary Computation) useful for our research work. JCLEC is an open source, platform independent and implemented in java. We need to specify the GP parameters in the configuration file of JCLEC toolkit which is in XML file format, to run the experiments.

### (a) Configuration parameters

First of all we have to select an algorithm to solve the problem. We have selected SGE (Simple Generational and Elitist Algorithm) which is available in JCLEC specifically for genetic programming.

*<process    algorithm-type    ="net.sf.jclec.algorithm.classic. SGE">*

Standard GP uses tree representation to represent an individual. We have also used tree representation to represent an individual. Thus, the package *net.sf.jclec.exprtree* must be used, establishing the minimum tree size, the maximum tree size and the list of terminal symbols and functions. Below we present that how to set the tree size, terminals and functions.

*<species type ="net.sf.jclec.exprtree.ExprTreeIndividual - Species" >*

*<min-tree-size> 3 </min-tree-size>*
*<max-tree-size> 25 </max-tree-size>*

*<terminal class = " tutorial.symreg.X "/>*

*<function class="tutorial.symreg.Add"/>*
*<function class="tutorial.symreg.Sub"/>*
*<function class="tutorial.symreg.Mul"/>*

The population is randomly initialized by using expression trees and the class *net.sf.jclec.exprtree.ExprTreeCreator*.

*< provider type ="net.sf.jclec.exprtree.ExprTreeCreator"/>*

We need to specify the max of generation for stopping criterion.

*<max-of-generations> 100 </max-of-generations>*

Selection of parents can be set by using the *net.sf.jclec.selector* package. Tournament selection gives the better performance, so we have selected the tournament selector as parent selector.

*<parents-selectors type = "net.sf.jclec.selector. Tournament-Selector" tournament-size="7"/>*

Fitness function calculated using evaluator. The declaration of evaluator type is mandatory. We use the symbolic regression problem so we specified SymregEvaluator as evaluator type.

*<evaluator type="tutorial.symreg.SymregEvaluator"/>*

*(b)  packages used for Experiment:*

*net.sf.jclec.fitness* This package contains several implementations of the IFitness interface.

*net.sf.jclec.selector* This package has implementations for several selection methods. Boltzmann Selector, Random Selector, Roulette selector, Stochastic remaining selection, Universal stochastic selection, Range selection, Tournament selection are available selectors in this package.

*net.sf.jclec.exprtree* contains the ExprTreeIndividual which defines a type of individual. This package also contains the ExprTreeIndividualSpecies class that defines the structure of individuals and operators to manipulate them continuously. Subtree Crossover, Tree Crossover, AllNodesMutator, DemoteMutator, GrowMutator, OneNodeMutator, Promote-Mutator, PromoteMutator, TruncMutator are the available operators for GP in JCLEC.

*(c)  Implementation details:*

Multi-dimensional symbolic regression problem solving facility was not available in JCLEC. We have implemented it by modifying *SymregEvaluator* class file. There are only three functions are available, that is Addition, Multiplication and Subtraction. We implemented following functions *division, sin, cosine, tan, square root, exponential* and *log* in JCLEC for our experiments. To use the newly created terminals and functions we need to set them in configuration file.

We modified the *seed generator* class file to pass the current time as a seed, rather than the static seed. For the comparing of performance of our proposed crossover operator with standard subtree crossover and soft brood crossover, we have to generate the graphs of fitness versus generation. For that we have modified the *PopulationReporter* class file that generates the .csv file that contains the generation and fitness.

JCLEC does not have the support for soft brood crossover operator. So we have implemented it. Only subtree crossover and tree crossover operators are available in JCLEC for genetic programming. Subtree crossover operators performs the crossover with the branches of tree where as tree crossover performs the crossover with the whole tree. For the implementation of soft brood crossover operator we have modified the *SubtreeCrossover* class and *ExprTree-Recombinator* class files.

We modified the *SubtreeCrossover* class file because it contains the logic of crossover point selection and helpful to implement the proposed crossover. The modification of *ExprTreeRecombinator* class file is required because it contains the method that called the genetic operator which is set into the configuration file that is in xml format.

## IV EXPERIMENTS

We have performed the experiments on three different symbolic regression problems.

Problem 1: $cos\ (\sqrt{sin\ (q)}) * cos\ (p) * sin(x) + tan(r\text{-}s)$
Problem 2:  Sextic Polynomial: $x^6 - 2x^4 + x^2$
Problem 3: $2x^2\text{-}3x + 4$

TABLE III
GP PARAMETERS FOR THE ABOVE PROBLEMS

| Parameters | Value |
|---|---|
| Population size | 100 |
| Maximum Generation | 50 |
| Min Tree Size | 3 |
| Max Tree Size | 25 |
| Terminal Set for Problem 1 | {X,P,Q,R,S} |
| Terminal Set Problem 2 | {X} |
| Terminal Set Problem 3 | {X, Constants(0 to 1)} |
| Function Set for Problem 1 | {+, -, *, Sqrt, Sin, Cos, Tan} |
| Function Set for Problem 2 and Problem 3 | {+, -, *} |
| Parent selector | Tournament selector with size 7 |
| Crossover Probability | 0.8 |
| Mutation Probability | 0.1 |

For the Problem 1, 2 and 3 we have set the GP parameters as shown in the Table III. And we have prepared the results of 30 runs for each problem using subtree crossover, soft brood crossover and modified soft brood crossover operators.

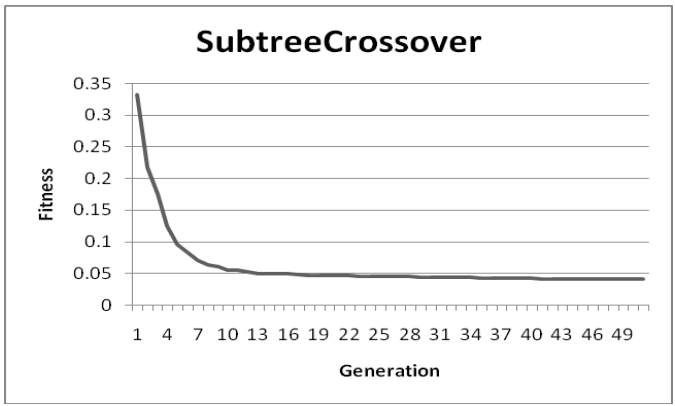

Fig. 1 Plot of Generations v/s Fitness for Problem1 using SubtreeCrossover

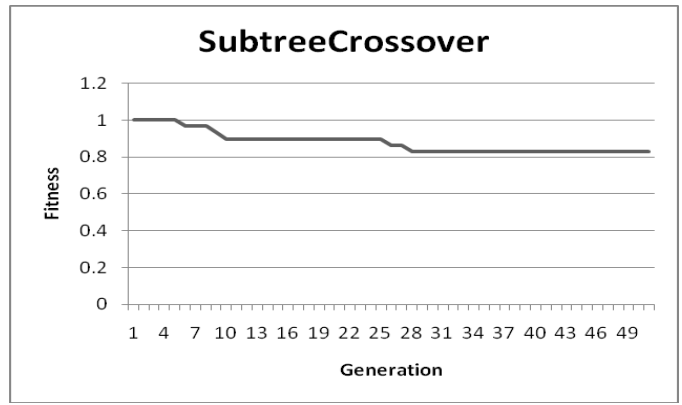

Fig. 4 Plot of Generations v/s Fitness for Problem2 using SubtreeCrossover

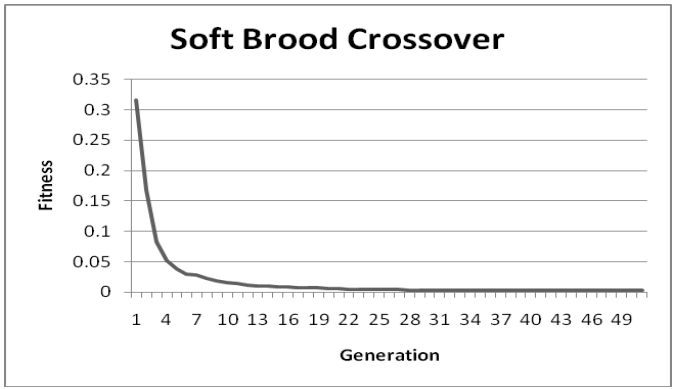

Fig. 2 Plot of Generations v/s Fitness for Problem 1 using Soft Brood Crossover

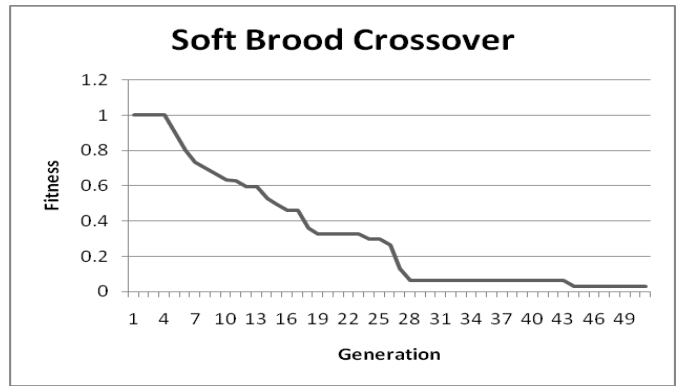

Fig. 5 Plot of Generations v/s Fitness for Problem2 using Soft Brood Crossover

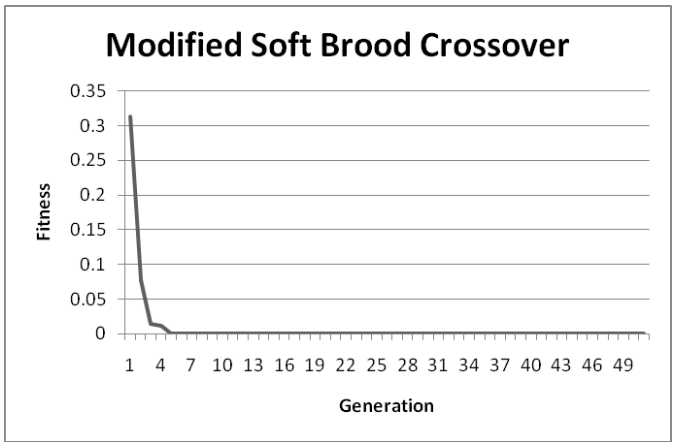

Fig. 3 Plot of Generations v/s Fitness for Problem1 using Modified Soft Brood Crossover.

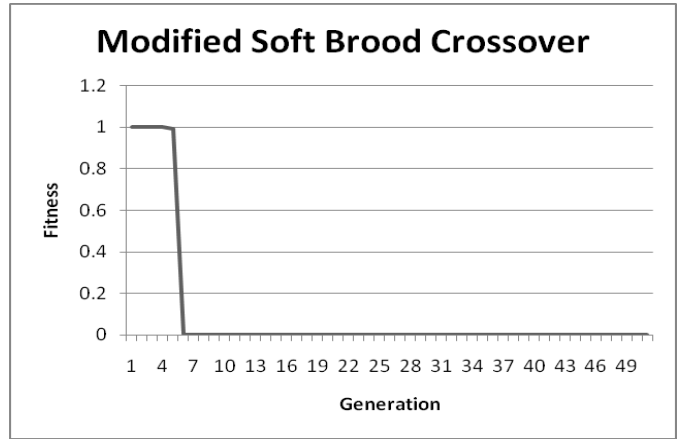

Fig. 6 Plot of Generations v/s Fitness for Problem2 using Modified Soft Brood Crossover

Figure 1 shows that best fitness obtained at 48th generation using subtree crossover. Figure 2 shows that best fitness obtained at 39th generation using Soft brood crossover and Figure 3 represents that best fitness obtained at 5th generation using Modified soft brood crossover. For the Problem 1, our proposed crossover gives the best fitness into less number of generations compare to subtree crossover and soft brood crossover operators.

Figure 4 shows that best fitness obtained at 27th generation using subtree crossover. Figure 5 shows that best fitness obtained at 43rd generation using Soft brood crossover and Figure 6 represents that modified soft brood crossover obtained best fitness at 5th generation. From the obtained results for Problem 2, we can say that our proposed crossover obtains the best fitness into less number of generations compare to soft brood crossover and subtree crossover operators.

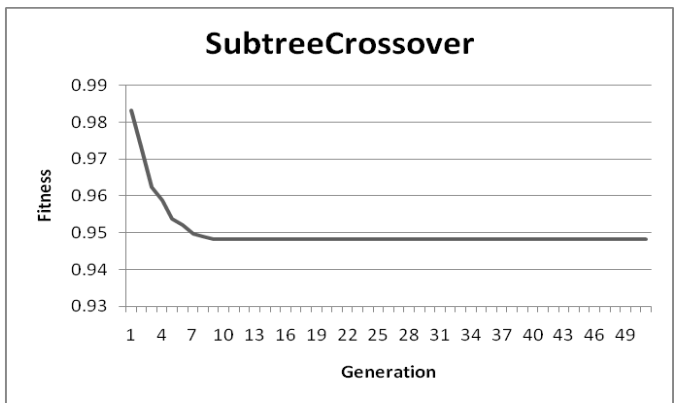

Fig. 7 Plot of Generations v/s Fitness for Problem3 using SubtreeCrossover

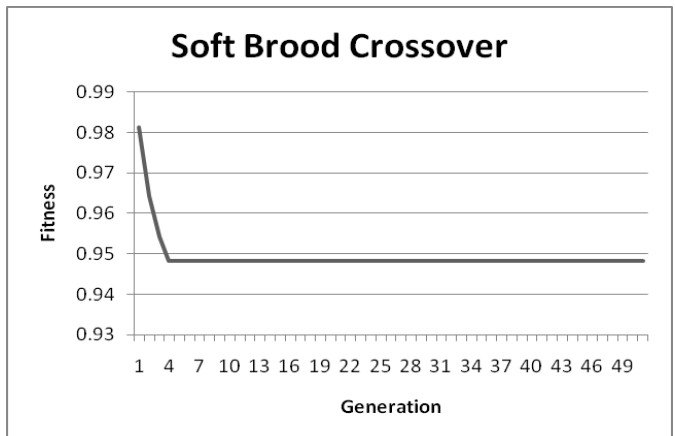

Fig. 8 Plot of Generations v/s Fitness for Problem3 using Soft Brood Crossover

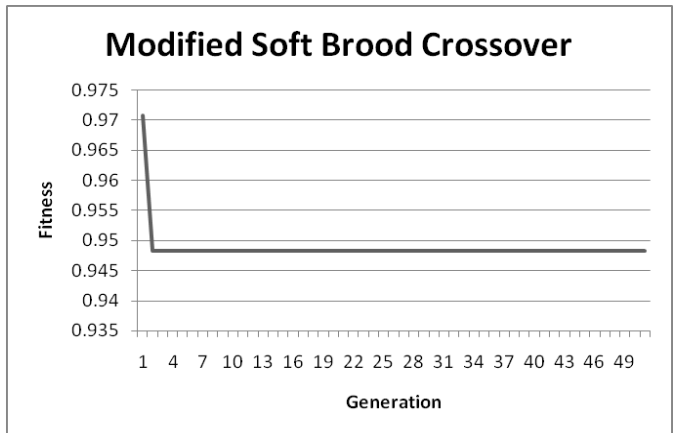

Fig. 9 Plot of Generations v/s Fitness for Problem3 using Modified Soft Brood Crossover

Figure 7 shows that subtree crossover obtained the best fitness at 8th generation. Figure 8 shows that Soft brood crossover obtained best fitness at 3rd generation and Figure 9 represents that modified soft brood crossover obtained best fitness at 2nd generation. In the case of Problem 3, modified soft brood crossover obtained the same fitness as soft brood crossover and subtree crossover. But, modified soft brood crossover obtained the fitness in less number of generations than the other two crossover operators.

## CONCLUSIONS

We proposed a new crossover operator for genetic programming that modifies the existing soft brood crossover operator. We have implemented soft brood crossover and proposed crossover (modified soft brood crossover) into the JCLEC toolkit. Then we have performed the experiments on three different symbolic regression problems (high dimension, sextic polynomial, symbolic regression with constants) using subtree crossover, soft brood crossover and modified soft brood crossover operators. From the obtained results for three different problems, we can conclude that our proposed crossover (Modified Soft Brood Crossover) gives good performance than the existing Soft Brood Crossover and Subtree Crossover operators.